\documentclass{article}

\usepackage{xcolor}
\def\maketitle{
\@author@finish
\title@column\titleblock@produce
\suppressfloats[t]}

\usepackage{times}
\usepackage[utf8]{inputenc} 
\usepackage[T1]{fontenc}    
\usepackage{url}            
\usepackage{booktabs}       
\usepackage{nicefrac}       
\usepackage{microtype}      
\usepackage{graphics,color}
\usepackage{stmaryrd}

\usepackage{algorithm}
\usepackage{algorithmic}
\usepackage{enumitem}

\usepackage{amsfonts}       
\usepackage{amsmath}       
\usepackage{amssymb}

\newcommand{\Fig}[1]{Figure~\ref{#1}}  
\newcommand{\fig}[1]{Fig.~\ref{#1}}    

\newcommand{\tab}[1]{Table~\ref{#1}}

\newcommand{\eqn}[1]{Eq.~\ref{#1}} 
\renewcommand{\sec}[1]{Sec.~\ref{#1}} 
\newcommand{\supp}[1]{Suppl.~\ref{#1}}

\newcommand{\figref}[1]{\fig{#1}}    

\newcommand{\tabref}[1]{\tab{#1}}

\newcommand{\secref}[1]{\sec{#1}} 

\usepackage{xspace}
\makeatletter
\DeclareRobustCommand\onedot{\futurelet\@let@token\@onedot}
\def\@onedot{\ifx\@let@token.\else.\null\fi\xspace}

\def\ie{i.e\onedot}

\makeatother

\DeclareMathOperator*{\argmin}{arg\,min}

\definecolor{ourblue}{rgb}{0.368,0.507,0.71}
\definecolor{ourorange}{rgb}{0.881,0.611,0.142}
\definecolor{ourgreen}{rgb}{0.56,0.692,0.195}
\definecolor{ourred}{rgb}{0.923,0.386,0.209}
\definecolor{ourviolet}{rgb}{0.528,0.471,0.701}
\definecolor{ourbrown}{rgb}{0.772,0.432,0.102}
\definecolor{ourlightblue}{rgb}{0.364,0.619,0.782}
\definecolor{ourdarkgreen}{rgb}{0.572,0.586,0.}

\definecolor{ourcyan2}{rgb}{0.125,0.722,0.804}
\definecolor{ourred2}{rgb}{0.863,0.184,0.047}
\definecolor{ouryellow2}{cmyk}{0,0.16,1.0,0.07}
\definecolor{ourviolet2}{cmyk}{0.55,0.56,0,0.47}
\definecolor{ourorange2}{cmyk}{0,0.46,0.89,0.11}

\usepackage{amsthm}
\usepackage{graphicx}
\usepackage{mathtools}
\usepackage{caption}
\usepackage{subcaption}
\usepackage{wrapfig}
\usepackage{xspace}
\usepackage{enumitem}

\usepackage{xr-hyper}
\makeatletter
\newcommand*{\addFileDependency}[1]{
  \typeout{(#1)}
  \@addtofilelist{#1}
  \IfFileExists{#1}{}{\typeout{No file #1.}}
}
\makeatother
\newcommand*{\myexternaldocument}[1]{%
    \externaldocument{#1}%
    \addFileDependency{#1.tex}%
    \addFileDependency{#1.aux}%
  }
  
\def\papertitle{Learning Agile Skills via Adversarial Imitation of Rough Partial Demonstrations}

\usepackage[final]{corl_2022} 

\title{\papertitle{}}

\usepackage{amsmath,amsfonts,bm}

\newcommand{\ddtw}{d^{\mathrm{DTW}}}
\newcommand{\traj}{\tau}

\newcommand{\solo}{Solo 8\xspace}
\newcommand{\method}{WASABI\xspace}
\newcommand{\baseline}{LSGAN\xspace}

\newcommand{\piref}{\gM}
\newcommand{\wD}{w^\mathrm{D}}
\newcommand{\wGP}{w^\mathrm{GP}}
\newcommand{\wI}{w^\mathrm{I}}
\newcommand{\rI}{r^\mathrm{I}}
\newcommand{\rT}{r^\mathrm{T}}
\newcommand{\rR}{r^\mathrm{R}}

\def\solowave{\textsc{SoloWave}}
\def\solobf{\textsc{SoloBackFlip}}
\def\solobfhcf{\solobf$^*$}
\def\solobfhcp{\textsc{\solobf$^\dagger$}}

\def\sololeap{\textsc{SoloLeap}}
\def\solostandup{\textsc{SoloStandUp}}
\def\solostanduphcf{\solostandup$^*$}
\def\solostanduphcp{\textsc{\solostandup$^\dagger$}}

\def\anymalwave{\textsc{ANYmalWave}}
\def\anymalbf{\textsc{ANYmalBackFlip}}

\newcommand{\ourlegend}{
{\small {{\color{ourgreen} \rule[.5ex]{2em}{1.5pt}} \method{}$^*$  \qquad
\textcolor{ourgreen}{\hbox to 20pt{\leaders\hbox to 4pt{\hss \textbf{-} \hss}\hfil}} \method{}$^\dagger$ } \qquad
{\color{ourred}\rule[.5ex]{2em}{1.5pt}} \baseline{}$^*$ \qquad
\textcolor{ourred}{\hbox to 20pt{\leaders\hbox to 4pt{\hss \textbf{-} \hss}\hfil}} \baseline{}$^\dagger$}
} 

\newcommand{\ourlegendvert}{
{\small 
\begin{tabular}{l@{\ }l}
{\color{ourgreen} \rule[.5ex]{2em}{1.5pt}} & \method{}$^*$  \\[.1em]
\textcolor{ourgreen}{\hbox to 20pt{\leaders\hbox to 4pt{\hss \textbf{-} \hss}\hfil}} &\method{}$^\dagger$ \\[.1em]
{\color{ourred}\rule[.5ex]{2em}{1.5pt}} & \baseline{}$^*$ \\[.1em]
\textcolor{ourred}{\hbox to 20pt{\leaders\hbox to 4pt{\hss \textbf{-} \hss}\hfil}} & \baseline{}$^\dagger$
\end{tabular}
}
} 

\newcommand{\ourlegendexp}{
{\small {{\color{ourgreen} \rule[.5ex]{0.5em}{2.5pt}} \method  \quad \color{ourred}\rule[.5ex]{0.5em}{2.5pt}} \baseline{}}\\[0.25em]}

\def\1{\bm{1}}

\DeclareMathAlphabet{\mathsfit}{\encodingdefault}{\sfdefault}{m}{sl}
\SetMathAlphabet{\mathsfit}{bold}{\encodingdefault}{\sfdefault}{bx}{n}

\def\gC{{\mathcal{C}}}

\def\gL{{\mathcal{L}}}
\def\gM{{\mathcal{M}}}

\def\gO{{\mathcal{O}}}

\def\gS{{\mathcal{S}}}
\def\gT{{\mathcal{T}}}

\def\sR{{\mathbb{R}}}

\newcommand{\E}{\mathbb{E}}

\usepackage[utf8]{inputenc}
\usepackage{pgfplots}
\DeclareUnicodeCharacter{2212}{−}
\usepgfplotslibrary{groupplots,dateplot}
\usetikzlibrary{patterns,shapes.arrows}
\pgfplotsset{compat=newest}

\usepackage[
	textsize=tiny,
	textwidth=2.2cm
	]{todonotes}

\makeatletter

\newcommand{\mvt}[2][]{\todo[backgroundcolor=blue!30!white, linecolor=blue!50!white, #1]{Marin: #2}}
\newcommand\@mvi[2]{\ifx\nocomments\undefined \textcolor{blue}{[Marin: #2]}\fi}
\def\mv{\@ifnextchar*\@mvi\mvt}

\author{
  Chenhao Li$^{1,2}$, Marin Vlastelica$^1$, Sebastian Blaes$^1$, Jonas Frey$^{2,1}$, \and \textbf{Felix Grimminger}$^1$, \textbf{Georg Martius}$^1$
  \and
  {\normalfont\normalsize $^{1}$Max Planck Institute for Intelligent Systems, Germany}
  \and
  {\normalfont\normalsize$^{2}$ Robotic Systems Lab, ETH Zurich, Switzerland}
  \\
  {\normalfont\normalsize\texttt{chenhao.li@tuebingen.mpg.de}}
}
\myexternaldocument{suppl}

\begin{document}

\maketitle


\begin{abstract}
    Learning agile skills is one of the main challenges in robotics.
    To this end, reinforcement learning approaches have achieved impressive results.
    These methods require explicit task information in terms of a reward function or an expert that can be queried in simulation to provide a target control output, which limits their applicability.
    In this work, we propose a generative adversarial method for inferring reward functions from partial and potentially physically incompatible demonstrations for successful skill acquirement where reference or expert demonstrations are not easily accessible.
    Moreover, we show that by using a Wasserstein GAN formulation and transitions from demonstrations with rough and partial information as input, we are able to extract policies that are robust and capable of imitating demonstrated behaviors.
    Finally, the obtained skills such as a backflip are tested on an agile quadruped robot called \solo{} and present faithful replication of hand-held human demonstrations.
\end{abstract}


\begin{figure}[h]
    \centering
    \begin{subfigure}[t]{.55\linewidth}
        \includegraphics[width=\linewidth]{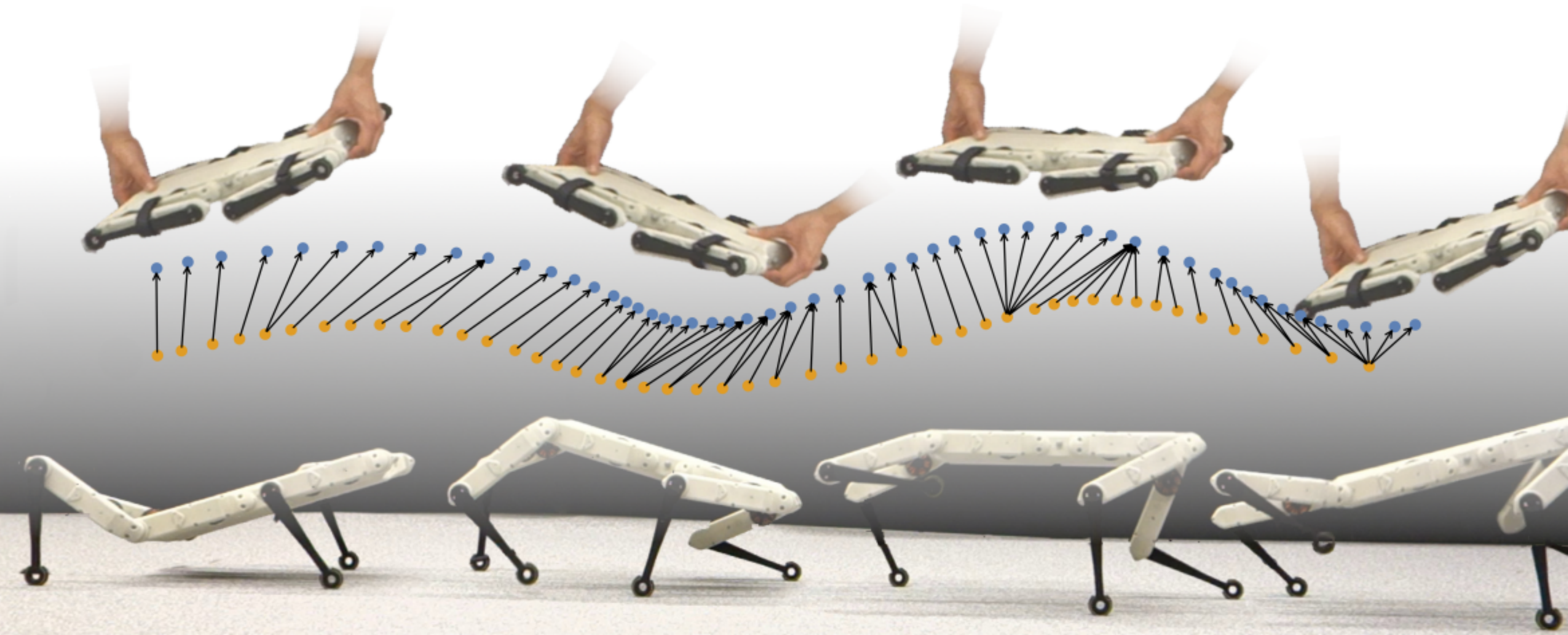}
    \end{subfigure}
    \begin{subfigure}[t]{.33\linewidth}
        \includegraphics[width=\linewidth]{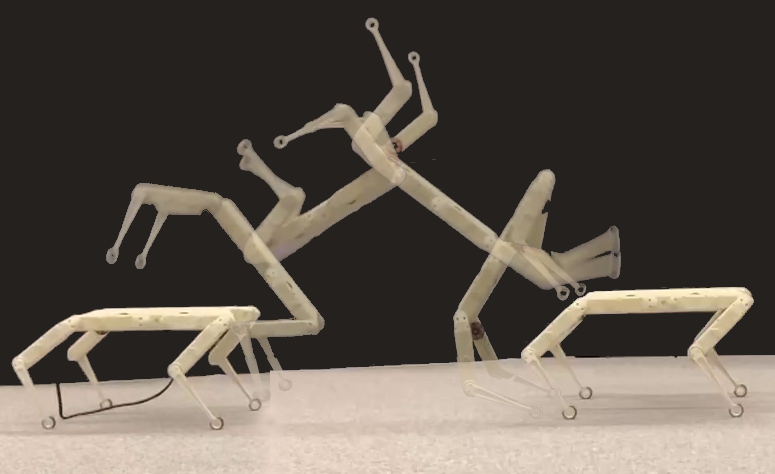}
    \end{subfigure}
    \caption{Our method (\method) achieves agile physical behaviors from rough (hand-held) and partial (robot base) motions. The illustrated performance measure is the Dynamic Time Warping distance of the base trajectories (left). A learned backflip policy is deployed on \solo{} (right).}
    \label{fig:demo-distance}
\end{figure}

\section{Introduction}
	
    Obtaining dynamic skills for autonomous machines has been a cardinal challenge in robotics. 
    In the field of legged systems, many attempts have been made to attain diverse skills using conventional inverse kinematics techniques~\citep{raibert2008bigdog, di2018dynamic}. 
    In recent years, learning-based quadrupedal locomotion has been achieved by reinforcement learning (RL) approaches to address more complex environments and improve performance~\citep{hwangbo2019learning, lee2020learning, kumar2021rma, miki2022learning}. 
    However, the demand for acquiring more highly dynamic motions has brought new challenges to robot learning. 
    A primary shortage of motivating desired behaviors by reward engineering is the arduous reward-shaping process involved. It can sometimes become extremely demanding in developing highly dynamic skills such as jumping and backflipping, where various terms of motivation and regularization require elaborated refinement.
    
    Given the availability of some expert references, one possible solution is Imitation Learning (IL), which aims to mimic expert behaviors in a given task. In this framework, the agent is trained to perform a task from demonstrations by learning a mapping between observations and actions with either offline (e.g. behavioral cloning~\citep{pomerleau1991efficient, torabi2018behavioral}) or interactive (e.g. DAgger, SMILe~\citep{ross2011reduction}) methods. 
    Generic IL methods could potentially reduce the problem of teaching a task to that of providing demonstrations, without the need for explicit programming or designing reward functions specific to the task~\citep{hussein2017imitation}. Another related approach to replicating exerted motions of an expert is Inverse Reinforcement Learning (IRL). 
    In IRL, the expert reward function is inferred given its policy or observed behaviors~\citep{ziebart2008maximum, bashir2021inverse, garg2021iq}. IRL is in general computationally expensive, and efforts are required to deal with ambiguous reward functions without making strong assumptions~\citep{abbeel2004apprenticeship}.
    
    More recently, Generative Adversarial Imitation Learning (GAIL)~\citep{ho2016generative} draws a connection between IL and generative adversarial networks (GANs)~\citep{goodfellow2014generative}, which train a generative model (generator) by having it deceive a discriminative classifier (discriminator).
    The task of the discriminator is to distinguish between data generated by the generator and the true data distribution. 
    In the setting of GAIL, the true data distribution is the expert state-action distribution, while the learned policy is treated as the generator.
    The output of the discriminator can then be used as a reward that encourages the learning agent to generate similar behaviors to the demonstration.
    Analogously, the technique has been used for learning adversarial motion priors (AMP)~\citep{peng2021amp}, where the output of the discriminator is used as an additional style reward to the actual task reward, that is available beforehand.
    In a sense, AMP enables solving well-defined tasks in a specific style specified by a reference motion, without requiring access to underlying expert actions. 
    
    In this work, we present a novel adversarial imitation learning method named Wasserstein Adversarial Behavior Imitation (\method{}). We show that we are able to extract sensible task rewards from rough and partial demonstrations by utilizing adversarial training for obtaining agile skills in a sim-to-real setting.
    In contrast to~\citet{peng2021amp}, our approach does not require any prior information about the task at hand in form of a specific reward function, but only reasonable task-agnostic regularization terms in addition to the adversarial reward that make the robot motion more stable.
    Most importantly, we achieve this without having access to samples from an expert policy, but rather hand-held human demonstrations that are physically incompatible with the robot itself. To the best of our knowledge, this is the first time that highly dynamic skills are obtained from limited reference information.
    In summary, our contributions include: 
    \textbf{(i)} An adversarial approach for learning from partial, physically incompatible demonstrations. 
    \textbf{(ii)} Analysis of the Least-Squares vs. Wasserstein GAN loss for reward inference.
    \textbf{(iii)} Experimental validation in simulation and on a quadruped robot.
    Supplementary videos for this work are available at \url{https://sites.google.com/view/corl2022-wasabi/home}.
    

\section{Related Work}
	
	Advances in robotics have spawned many potential applications that require intelligent systems to be able to not only make decisions but also to perform physical movements expectedly.
	However, in many cases, the desired behavior may not be discovered by a learning agent due to sub-optimal parameter settings or algorithmic limitations~\citep{pathak2019self, cai2020provably}.
	While learning a task might be stated as an optimization problem, it has become widely accepted that having prior knowledge provided by an expert is more effective and efficient than attempting to solve the problem from scratch~\citep{billard2008survey, argall2009survey}.
	
	The idea of IL has been formed decades ago, raising solutions in conceptual and computational models to replicate motions from demonstrations~\citep{schaal1996learning, atkeson1997robot, schaal1999imitation}.
	It has been commonly acknowledged that IL entails three major approaches: model-based IL, learning a control policy directly, and learning from demonstrated trajectories.
	In the first approach, algorithms are applied to learn the parameters of the dynamics model to ensure that all executed motions closely follow the demonstration~\citep{calinon2010learning, khansari2011learning, ijspeert2013dynamical}.
	In the second approach, also known as behavioral cloning, the agent tries to reproduce the observed state-action pairs of the expert policy~\citep{pomerleau1991efficient, torabi2018behavioral}.
	Behavioral cloning often faces the problems of error compounding and poor generalization, which can lead to unstable policy output, particularly in out-of-distribution regions~\citep{codevilla2019exploring}.
	Alternatively, reference motions can be learned using an imitation goal, which is often implemented as a tracking objective that aims to reduce the pose error between the simulated character and target poses from a reference motion~\citep{lee2010data, liu2010sampling, liu2016guided, peng2018deepmimic}.
	A common strategy to estimate the pose error is to use a phase variable as an additional input to the controller to synchronize the agent with a specific reference motion~\citep{lee2019scalable, peng2018deepmimic, peng2018sfv}.
	This method typically works well for replicating single motion clips, but it may fail to scale to datasets with multiple reference motions which may not be synchronized and aligned according to a single-phase variable~\citep{peng2021amp}.

	Instead of employing a handcrafted imitation objective, adversarial IL techniques train an adversarial discriminator to distinguish between behaviors generated by an agent and demonstrations~\citep{abbeel2004apprenticeship, ho2016generative}.
	While these methods have shown some promise for motion imitation tasks~\citep{merel2017learning, wang2017robust}, adversarial learning algorithms are notoriously unstable, and the resulting motion quality still lags well behind that of state-of-the-art tracking-based systems.
	Especially in the low-data regime, adversarial models can take a long time to converge~\citep{jena2020loss, peng2021amp}.
	In some cases, adversarial IL techniques show limited robustness against different environment dynamics, as it fails to generalize to tasks where there is considerable variability in the environment from the demonstrations~\citep{fu2017learning}.

	With the ability to encompass multiple reference motions, AMP decouples task specification from style specification by combining GAIL with extra task objectives~\citep{peng2021amp, escontrela2022adversarial}.
	The use of AMP reduces efforts in the selection of distance error metrics, phase indicators, and appropriate motion clips. 
	This allows the learning agent to execute tasks that may not be portrayed in the original demonstrations.
	To enable active style control, Multi-AMP allows for the switching of multiple different style rewards by training multiple discriminators encoding different reference motions in parallel~\citep{vollenweider2022advanced}.
    

\section{Approach}
	
    In this section, we describe our method, \method{}, which involves generative adversarial learning of an imitation reward from rough and partial demonstrations using a GAN framework.
    
    \subsection{Learning Task Reward from Limited Demonstration Information}\label{sec:wgan}

    We consider \emph{partial} demonstrations that are given in terms of limited state observations, for instance only local velocities of the robot's base. 
    The demonstrations are formulated as sequences of $o_t\in \gO$, where the full state space $\gS$ of the underlying Markov Decision Process can be mapped to the observation space $\gO$ with a function $\Phi: \gS \to \gO$. 
    We utilize generative adversarial learning for inferring the task reward function from such demonstrated transitions $(o, o')$ in a reference motion.
    As such, the discriminator in this setup is to distinguish samples of the policy transition distribution $d^\pi$ from the reference motion distribution $d^\piref$.
    The policy $\pi$ takes on the role of the generator. 

    The original GAN min-max loss (CEGAN) formulation has shown to suffer from vanishing gradients due to saturation regions of the cross-entropy loss function which slows down training~\citep{arjovsky2017towards}.
    If the discriminator performs excessively well and thus becomes saturated, the policy will not be able to learn any information, since it receives a constant penalty for being far away from the demonstrations.
    For this reason, \citet{peng2021amp} propose to use the least-squares GAN (LSGAN) loss~\citep{mao2017least} in AMP as a substitute for reward function learning.
    The LSGAN loss is formulated as
    \begin{equation}
        \argmin_D \E_{d^{\mathcal{M}}} \left [ \big(D(o, o') - 1 \big)^2 \right ] + \mathbb{E}_{ d^{\pi}} \left [ \big(D(\Phi(s), \Phi(s')) + 1 \big)^2 \right ].
    \end{equation}
    The discriminator is defined as a mapping $D: \gO \times \gO \mapsto \sR$ and can be used, together with $\Phi$, as a drop-in replacement for the unknown reward function $r(s,s')$.
    Intuitively, the LSGAN loss forces the discriminator to output $+1$ for samples from the reference motion and $-1$ for those from the policy.
    It not only prevents vanishing gradients but also provides a well-scaled output that eases downstream policy learning.
    However, when faced with demonstrations that initially seem beyond what the agent can achieve, the discriminator is prone to be driven to optimality, prohibiting a more fine-grained evaluation of the policy transitions with respect to their closeness to the reference motion.
    Moreover, the LSGAN discriminator output does not directly lead to a practical reward function by itself, since an increase in its value does not always represent close replications of demonstrated transitions.
    This is a consequence of the least-squares loss symmetricity around $-1$ and $+1$, therefore a suitable mapping is typically needed to transform the output into a well-behaved reward function.
    For this reason, we propose to use the Wasserstein loss
    \begin{equation}\label{eqn:wasserstein}
        \argmin_D - \mathbb{E}_{d^{\piref}} \left [ D(o, o') \right ] + \mathbb{E}_{d^{\pi}} \left [ D(\Phi(s), \Phi(s')) \right ],
    \end{equation}
    especially for highly dynamic motions where the discriminator is more likely to optimally distinguish between the reference and the generated motions.
    Under conditions of Lipschitz continuity, the Wasserstein loss is an efficient approximation to the earth mover's distance which effectively measures the distance between two probability distributions~\citep{rubner1998metric}.
    In the original Wasserstein GAN (WGAN), \citet{Arjovsky2017WassersteinG} enforce Lipschitz continuity by projected gradient descent, \ie clipping the network weights. Similarly, we apply $L_2$ regularization on the discriminator for the sake of simplicity.
    In addition, discriminator weight regularization also controls the scale of its output, which results in stable imitation rewards.

    \subsection{Preventing Mode Collapse in Adversarial Reward Learning} \label{sec:preventing_mode_collapse}
    
    Mode collapse is a common problem in GAN training, which manifests itself by the generator being able to produce only a small set of outputs.
    In our framework, mode collapse is reflected by the policy trying to replicate only a subset of the reference motion which gives a high reward.
    
    The Wasserstein loss can alleviate mode collapse by allowing training of the discriminator to optimality while avoiding vanishing gradients~\citep{Arjovsky2017WassersteinG}.
    In fact, if the discriminator does not get stuck in the local minimum, it learns to reject partial behaviors on which the policy stabilizes.
    As a result, the policy will have to attempt something different, if possible. 
    In addition to the implementation of the Wasserstein loss, we extend the capability of the discriminator by allowing more than one state transition as input, \ie we extend the input to $H$ consecutive observations. 
    Note that this is typically not applicable to CEGAN or LSGAN, as a longer horizon makes the discriminator even stronger.
    By taking more sequential states into account, the policy reduces its chance to resort to the same safe transition patterns that are present in the reference motion.
    
    We denote trajectory segments of length $H$ preceding time $t$ by $o_t^H=(o_{t-H+1},\dots,o_{t})$
    for the reference observations and $s_t^H=(s_{t-H+1},\dots,s_{t})$ for the states induced by the policy. For clarity, we omit the time index in the following. 
    To simplify notation, we write $\Phi(s^H)$ to express that each state in $s^H$ is mapped to $\gO$.   
    In our experiments, we select linear and angular velocities $v$, $\omega$ of the robot base in the robot frame, measurement of the gravity vector in the robot frame $g$, and the base height $z$ as the observation space $\gO$.
    More information on the state space and demonstration space is detailed in \supp{app:sec:model_representation}.
    Note that in this example, no joint information is required by the discriminator.
    This facilitates the process to obtain the expert motion, as one 
    can simply move the robot base by hand along the desired trajectory without any joint actuation.
    
    Using $H$-step inputs and a gradient penalty, \eqn{eqn:wasserstein} turns into
    \begin{equation}
        \argmin_D \wD \left ( - \E_{d^{\piref}} \left [ D \left (o^H \right) \right ] + \E_{d^{\pi}} \left [ D \left (\Phi(s^H) \right ) \right ] \right ) + \wGP \E_{d^{\piref}} \left [ \big \|\nabla_\Omega D\left (\Omega \right )\mid_{\Omega=o^H)} \big \|_2^2 \right ],
        \label{eqn:discriminator_objective}
    \end{equation}
    where the last term denotes the penalty for nonzero gradients on samples from the dataset~\citep{peng2021amp}.
    $\wD$ and $\wGP$ denote the weights on the Wasserstein loss and the gradient penalty, respectively. In our experiments, they are set to $\wD = 0.5$ and $\wGP = 5.0$ for all tasks.
    
    \subsection{Reward Formulation} \label{sec:reward_normalization_and_termination_penalty}
    
    Despite discriminator regularization, due to the unbounded discriminator output, the scale of the reward can be arbitrary which makes it difficult to introduce additional regularization terms for stabilizing the robot motion.
    Therefore, we normalize the reward to have zero mean and unit variance in the policy training loop by maintaining its running mean $\widehat\mu$ and variance $\widehat{\sigma}^2$. 
    With this formulation, the imitation reward is then given by
    \begin{equation}
        \rI = \frac{D\left (\Phi(s^H) \right)-\widehat\mu}{\widehat{\sigma}},
        \label{eqn:imitation_reward}
    \end{equation}
    where  $D \left (\Phi(s^H) \right)$ denotes the output of the discriminator.
    
    To increase policy learning efficiency, a common practice is to define a termination condition for rollouts.
    In our work, an instantaneous environment reset is triggered when a robot base collision against the ground is detected.
    Since the imitation reward has zero mean and difficult behaviors are likely to result in negative rewards initially, 
    the policy may attempt to end the episode early.
    To circumvent this, a termination penalty is imposed at the last transition before a collision happens. 
    As the normalized reward follows a distribution with zero mean and unit variance, $-5 \sigma$ is a lower bound on the reward with a probability greater than $99.99\%$.
    We use this to derive a reasonable termination penalty, based on the geometric series, by a high-probability lower bound on the return
    \begin{equation}
        \rT = \llbracket s\in \gT\rrbracket \dfrac{-5\sigma}{1 - \gamma}, \label{eqn:termination}
    \end{equation}
    where $\gamma$ is the discount factor, $\gT$ is the set of early termination states, and $\llbracket \cdot \rrbracket$ is the Iverson bracket ($1$ if true, $0$ otherwise).
    Putting everything together, the total reward that the policy receives encompasses three parts, the imitation reward $\rI$ defined by the normalized discriminator output, the termination reward $\rT$, and the regularization reward $\rR$ to guarantee stable policy outputs (detailed in \supp{app:sec:regularization_reward_functions})
    \begin{equation}
        r = \wI (\rI  + \rT)+ \rR,
        \label{eqn:total_reward}
    \end{equation}
    where $\wI$ is a motion-specific scaling factor controlling the relative importance of the imitation reward  (and the termination penalty) with respect to the regularization terms.

\begin{figure}
        \centering
        \includegraphics[width=0.9\linewidth]{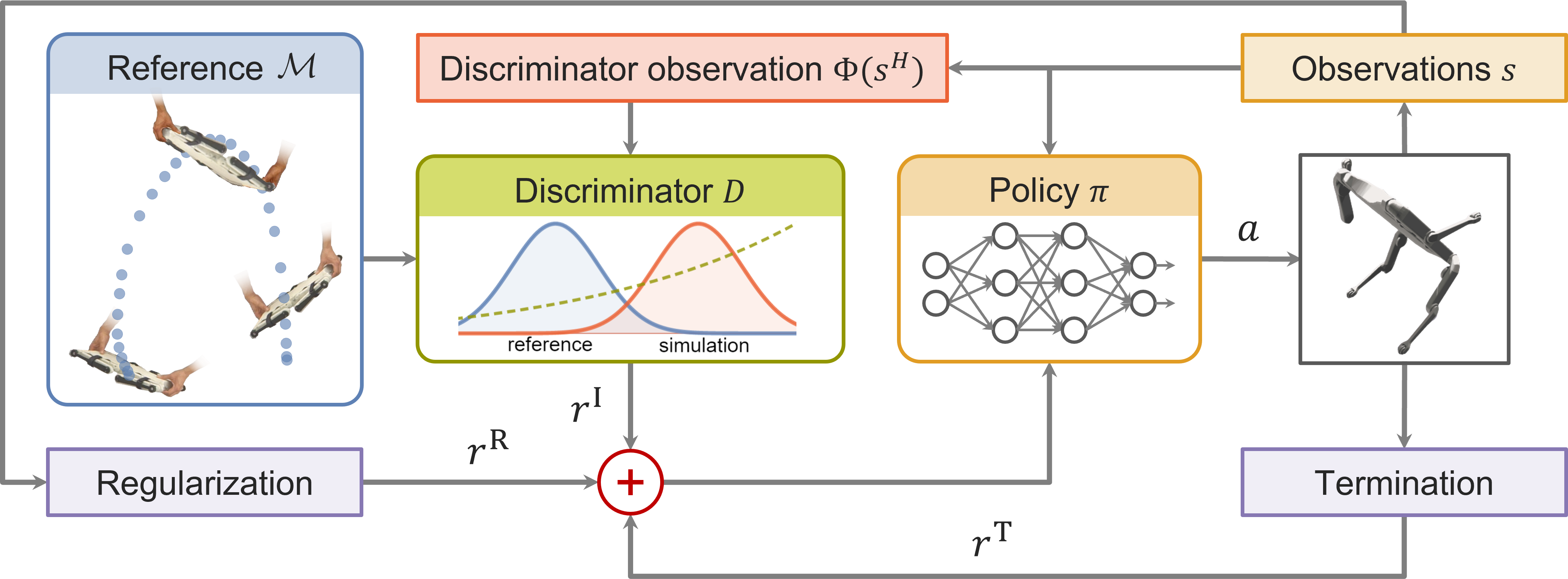}
        \caption{System overview. Given a reference dataset defining the desired base motion, the system trains a discriminator that learns an imitation reward for the policy training. This imitation reward is then combined with a regularization reward and termination penalty to train a policy that enables the robot to replicate the demonstrated motion while maintaining feasible and stable joint actuation.}
        \label{fig:system_overview}
    \end{figure}

    Note that our reward formulation enables the robot to learn highly dynamic skills without any explicitly defined desired-motion-incentivizing reward, as is used in AMP, where an a priori designed reward still has to motivate the policy to execute a specific movement~\citep{peng2021amp}.
    It is also noteworthy that the LSGAN formulation in our setting can be viewed as an implementation of AMP modified for task reward learning with substantial adaptations as detailed in \supp{app:sec:method_comparison}.
    \Fig{fig:system_overview} provides a schematic overview of our method, and an algorithm overview is detailed in Algorithm \ref{algorithm1}.

	\begin{algorithm}[htbp]
        \caption{\method}
        \footnotesize
        \label{algorithm1}
        \begin{algorithmic}[1]
            \STATE \textbf{Input}: dataset of reference motions $\mathcal{M}$, feature map $\Phi$
            \STATE initialize discriminator $D$, policy $\pi$, value function $V$, state transition buffer $s^H$, replay buffer $B$
            \FOR{learning iterations $ = 1,2,\dots$}
                    \STATE collect $N+H$  transitions $(s_t, a_t, \rR_t, s_{t+1}  )_{t-H}^{t+N}$ with policy $\pi$
                    \STATE compute $\rI_\tau$ using discriminator outputs $D \left(\Phi(s^H_i)\right)$ for $i=
                    t,\dots, t+N$
                    \STATE calculate transition rewards $r_t = \wI \left (\rI_t + \rT_t \right ) + \rR_t$ according to Equations~\ref{eqn:imitation_reward}, \ref{eqn:termination}, and \ref{eqn:total_reward}
                    \STATE fill replay buffer $B$ with $ \left (s_t, a_t, r_t, s_{t+1} , \Phi(s_t^H)\right )_t^{t+N}$
                \FOR{policy learning epoch $ = 1,2,\dots,n_\pi$}
                    \STATE sample transition mini-batches $b^\pi\sim B$
                    \STATE update $V$ and $\pi$ by PPO objective or another RL algorithm
                \ENDFOR
                \FOR{discriminator learning epoch $ = 1,2,\dots,n_D$} 
                    \STATE sample transition mini-batches $b^\pi\sim B$ and $b^\gM \sim \gM$ 
                    \STATE update discriminator $D$ using $b^\pi$ and $b^\mathcal{M}$ according to the loss associated with \eqn{eqn:discriminator_objective}
                \ENDFOR
            \ENDFOR
        \end{algorithmic}
    \end{algorithm}


\section{Experiments}\label{sec:experiments}

\begin{figure}
    \centering
    \begin{subfigure}[t]{.20\linewidth}
        \includegraphics[width=\linewidth]{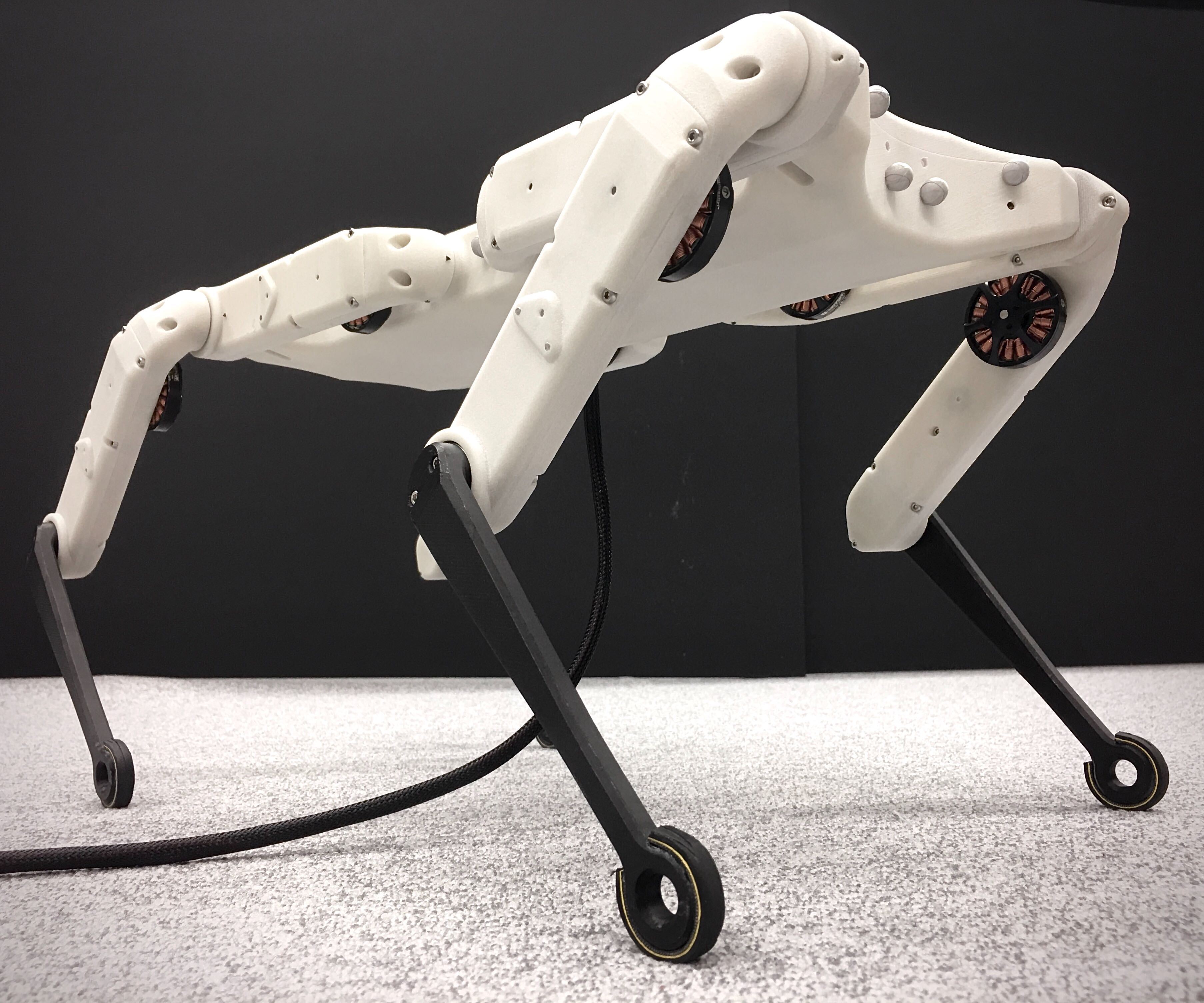}
    \end{subfigure}
    \begin{subfigure}[t]{.58\linewidth}
        \includegraphics[width=\linewidth]{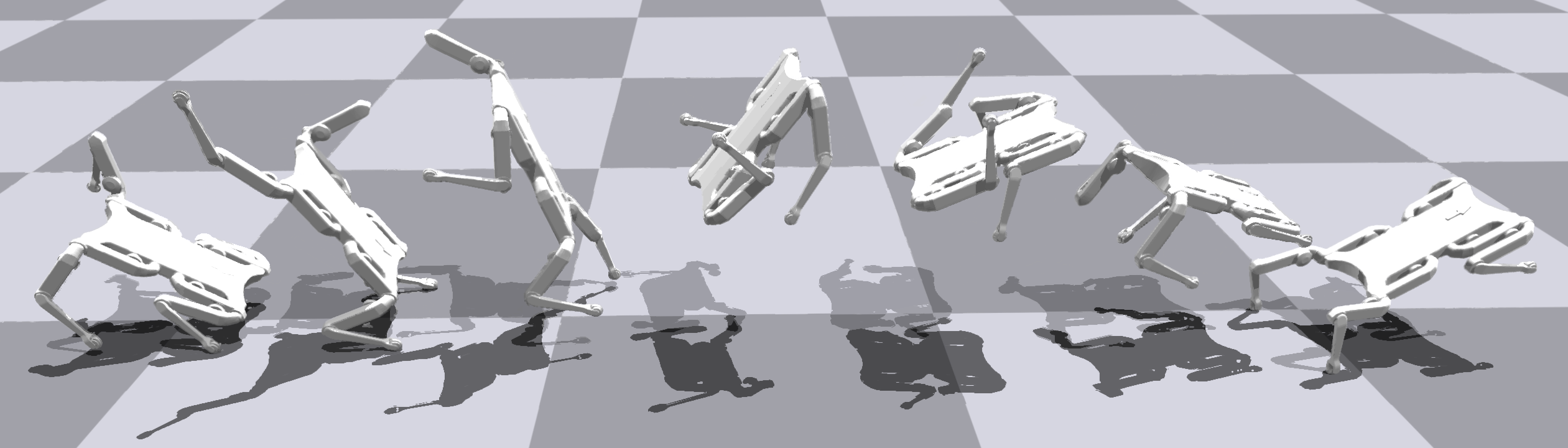}
    \end{subfigure}
	\caption{\solo{} (left). Backflip motion in Isaac Gym (right).}
    \label{fig:solo8}
\end{figure}

We evaluate \method{} on the \solo{} robot, an open-source research quadruped robot that performs a wide range of physical actions~\citep{grimminger2020open}, in simulation and on the real system (\figref{fig:solo8}).
For evaluation, we introduce 4 different robotics tasks.
In \sololeap{}, the robot is asked to move forward with a jumping motion. \solowave{} requires the robot to produce a wave-like locomotion behavior.  For \solostandup{} we require the robot to stand up on its hind legs. In \solobf{} the robot is asked to generate motions of a full backflip.
We provide \emph{rough} demonstrations of these motions by manually carrying the robot through the motion and recording only the base information.
The demonstrations are then used to infer an adversarial imitation reward for training a control policy that outputs target joint positions, as outlined in \sec{sec:wgan}.
An overview of the desired movements is provided in \supp{app:sec:tasks}, we also provide further ablation studies in \supp{app:sec:ablation}.

In all of our experiments, we use Proximal Policy Optimization (PPO)~\citep{schulman2017proximal} in Isaac Gym~\citep{makoviychuk2021isaac} and make use of domain randomization~\citep{tobin2017domain} for sim-to-real transfer.
Further details on the training procedure can be found in \supp{app:sec:training_details}.

\subsection{Induced Imitation Reward Distributions} \label{sec:wgan_vs_lsgan}

The LSGAN loss is proposed to alleviate the saturation problem that is encountered for the CEGAN loss.
Yet, as outlined in \secref{sec:wgan}, it does not directly yield a practical reward function.
\citet{peng2021amp} remedy this by using $\rI = \mathrm{max}\left[0,\;  1-0.25(D\left (\Phi(s), \Phi(s')\right) - 1)^2\right]$ to map the discriminator output to the imitation reward and bound it between $0$ and $1$.
However, with the effective clipping at $0$, information about the distance from the policy to the demonstration transitions is lost with discriminator prediction smaller than $-1$ (\figref{fig:reward-functions-hist}).
In addition, we show in \figref{fig:reward-functions-lsgan} that the imitation reward learned using \baseline{} yields a less informative signal for policy training, which is rather uniformly distributed across pitch rate $\dot{\theta}$ and base height $z$ dimensions.
In comparison, \method{} can use the discriminator output directly, learning a more characteristic reward function across the state space where reference trajectories are clearly outlined to yield high rewards in contrast to the off-trajectory states (\figref{fig:reward-functions-wasabi}).

\def\third{0.3\linewidth}
\def\half{0.49\linewidth}
\newlength{\figh}
\setlength{\figh}{0.23\linewidth}
\begin{figure}
    \centering

    \begin{subfigure}[b]{0.33\linewidth}
    \centering
        \includegraphics[height=\figh]{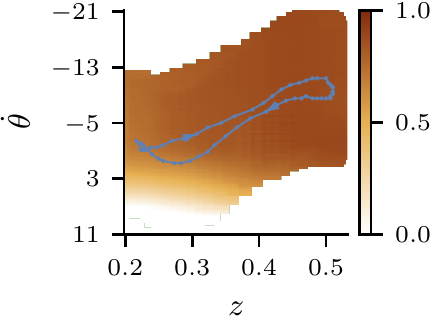}\vspace{-.3em}
        \caption{LSGAN}
        \label{fig:reward-functions-lsgan}
    \end{subfigure}\hspace{0.1em}
    \begin{subfigure}[b]{0.33\linewidth}
    \centering    
        \includegraphics[height=\figh]{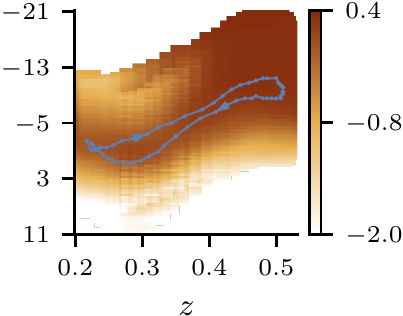}\vspace{-.3em}
        \caption{WASABI (WGAN)}
        \label{fig:reward-functions-wasabi}
    \end{subfigure}\hspace{0.1em}
    \begin{subfigure}[b]{0.29\linewidth}
        \centering
        \ourlegendexp
        \includegraphics[height=\figh]{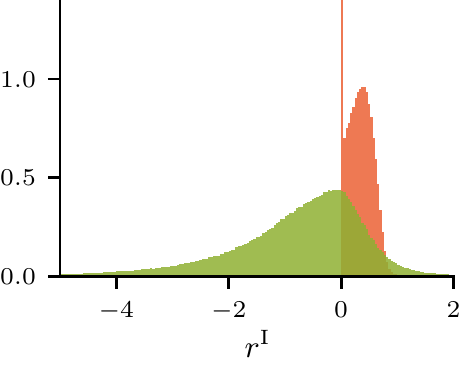}\vspace{-.3em}
        \caption{Distribution of $\rI$}
        \label{fig:reward-functions-hist}
    \end{subfigure}\vspace{-.5em}
    \caption{Adversarial imitation rewards for \solobf{}. Imitation reward heatmap for \baseline{} (a) and \method{} (b) around reference trajectories (blue) generated in varying pitch rate $\dot{\theta}$ and base height $z$. (c) Distribution of imitation rewards for \baseline{} and \method{}  during training. \method{} provides a more fine-grained reward function.}
    \label{fig:reward-functions}
\end{figure}

\subsection{Learning to Mimic Rough Demonstrations} \label{sec:learning_to_mimic_rough_demonstrations}

Since we record the base motion of the robot carried by a human demonstrator, we do not have access to a reward function evaluating learned behaviors or measuring the closeness between the demonstrated and the policy trajectories.
In addition, these trajectories are largely misaligned.
For this reason, we make use of Dynamic Time Warping (DTW)~\citep{berndt1994using} with the $L_2$ norm metric for comparing policy trajectories and reference demonstrations.
DTW allows us to match and compute the distance between the trajectories in a time-consistent manner (\fig{fig:demo-distance}). 
Concretely, we use $\E\left[\ddtw(\Phi(\traj_\pi), \traj_\gM)\right]$ as the evaluation metric, where $\traj_\pi\sim d^\pi$ is a state trajectory from a policy rollout and $\traj_\gM\sim d^\gM$ denotes a reference motion from the dataset.
We provide further details about this metric in \supp{app:sec:dtw}.
In \tabref{tab:performance} we compare performances in simulation for the different reference motions.

\begin{table}
    \centering
    \begin{tabular}{lllll} 
        \toprule
        Method  & \multicolumn{1}{c}{\sololeap} & \multicolumn{1}{c}{\solowave}  & \multicolumn{1}{c}{\solostandup}  &  \multicolumn{1}{c}{\solobf}   \\ 
        \midrule                                      
        \method{} &\boldmath$131.70\pm16.44$ & \boldmath$247.29\pm11.59$ & \boldmath $351.13\pm88.60$ &\boldmath $477.43\pm56.77$  \\
        \baseline{} & \boldmath$155.31\pm18.10$ & \boldmath $230.91\pm5.95$  &$678.21\pm6.71$ &$813.76\pm19.75$  \\
        \midrule
        Stand Still  & $216.41$  & $460.15$  &   $494.40$  & $877.74$    \\ 
        \bottomrule
    \end{tabular}
    \vspace{0.5em}
    \caption{Comparison of performances for \baseline{} and \method{} trained with hand-held demonstrations in terms of \textbf{DTW distance} $\ddtw$ (lower is better), successful runs are in \textbf{bold} font. As a reference, we provide also $\ddtw$ of a constantly standing trajectory.}
    \label{tab:performance}
\end{table}

\begin{figure}
    \centering 
    \vspace{1em}
    \begin{subfigure}[t]{.32\linewidth}
        \ \\[-1em]
        \includegraphics[width=\linewidth]{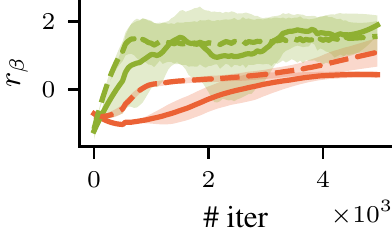}
    \end{subfigure}
    \hspace{1em}
    \begin{subfigure}[t]{.305\linewidth}
        \ \\[-1em]
        \includegraphics[width=\linewidth]{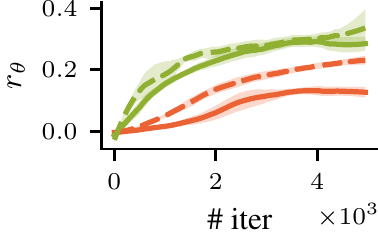}
    \end{subfigure}
    \hspace{1em}
    \begin{subfigure}[t]{.16\linewidth}
        \ \\
        \ourlegendvert
    \end{subfigure}
    
    \caption{Performance of \method{} and \baseline{} in terms of the handcrafted task reward for \solostandup{} (left) and \solobf{} (right). Dashed lines indicate partial information ($\dagger$).}
    \label{fig:wgan-vs-task-backflip}
\end{figure}

In order to confirm that \method{} is indeed able to extract a sensible reward function that motivates the desired motion, we compare the performance of \baseline{} and \method{} in \solostandup{} and \solobf{} using an expert baseline that is trained on a handcrafted task reward for generating demonstrations in simulation.
Details on the handcrafted task reward formulation are given in \supp{app:sec:handcrafted_task_reward}.
The learned policies are evaluated with the same task rewards that are used to obtain the expert policies. 
A comparison of training performance curves in terms of the corresponding handcrafted task rewards is detailed in \figref{fig:wgan-vs-task-backflip}.
In \tabref{tab:hc} we show the performance evaluation of the best runs.
Observe that the policies trained by \method{} perform comparably to the expert policies trained with the handcrafted rewards.
Interestingly, learning from partial state information may sometimes facilitate policy learning, since a decrease in discriminator observation dimensions could potentially alleviate the problem of discriminator becoming too strong as indicated in \figref{fig:wgan-vs-task-backflip}.

\begin{table}
    \centering
    \resizebox{0.98\linewidth}{!}{
    \begin{tabular}{lcccc} 
        \toprule
          Method & \solostanduphcp & \solostanduphcf & \solobfhcp & \solobfhcf  \\ 
        \midrule                                     
        \method{}       & \boldmath$1.54\pm0.51$ & \boldmath$1.68\pm0.51$  & \boldmath$0.36\pm0.05$  & \boldmath$0.28\pm0.02$ \\ 
        \baseline{}     & $1.07\pm0.5$     & $0.44\pm0.14$             &  $0.12 \pm 0.01$       & $0.06 \pm0.01$  \\ 
        \midrule
        Handcrafted &  \multicolumn{2}{c}{\boldmath$2.24 \pm 0.05$}    & \multicolumn{2}{c}{\boldmath$0.77 \pm 0.04$} 
        \\ 
        \bottomrule
    \end{tabular}}
    \vspace{0.5em}
    \caption{Performance comparison in terms of handcrafted \textbf{task reward} (higher is better). We denote with $*$ where the full robot configuration is given to the discriminator and $\dagger$ where only base information is given. Successful runs are in \textbf{bold} font. Std-dev.~is over 5 independent random seeds.}
    \label{tab:hc}
\end{table}

\subsection{Evaluation on Real Robot} \label{sec:evaluation_on_real_robot}

To evaluate our method on real system, we trained policies for sim-to-real transfer with \method{} for the \sololeap{}, \solowave{} and \solobf{}. 
The \solo{} robot is powered by an external battery and driven by a controller on an external operating machine.
It receives root state estimation using 10 markers attached around the base which are tracked using a Vicon motion capture system operating at 100\,Hz.
During deployment, we recorded the robot base information for evaluation by $\ddtw$.
As detailed in \supp{app:sec:sim-to-real_transfer}, the policy observation space, reward, and training hyperparameters are adapted to facilitate sim-to-real transfer for these tasks specifically.
The resulting performance on the real system, as shown in \tabref{tab:dtw_deployment}, resembles the performance obtained in simulation.

\begin{table}[htb]
    \centering
    \begin{tabular}{lccc} 
        \toprule
        & \sololeap & \solowave & \solobf \\ 
        \midrule
        \method (Real)  & $153.64 \pm \phantom{1}7.08$  & $215.38 \pm 21.82$ & $504.26 \pm 18.90$ \\ 
        \color{gray}\method{} (Sim) & \color{gray}$131.70\pm16.44$ & \color{gray}$247.29\pm11.59$  & \color{gray}$477.43\pm56.77$  \\
        \bottomrule
    \end{tabular}
    \vspace{0.5em}
    \caption{Sim-to-real performance on the \solo{} in terms of DTW distance (lower is better). Values are computed from the recorded data of the learned policies with respect to the reference trajectories.}
    \label{tab:dtw_deployment}
\end{table}

\subsection{Cross-platform Imitation}

\begin{figure}
    \centering
    \begin{subfigure}[t]{.15\linewidth}
        \includegraphics[width=\linewidth]{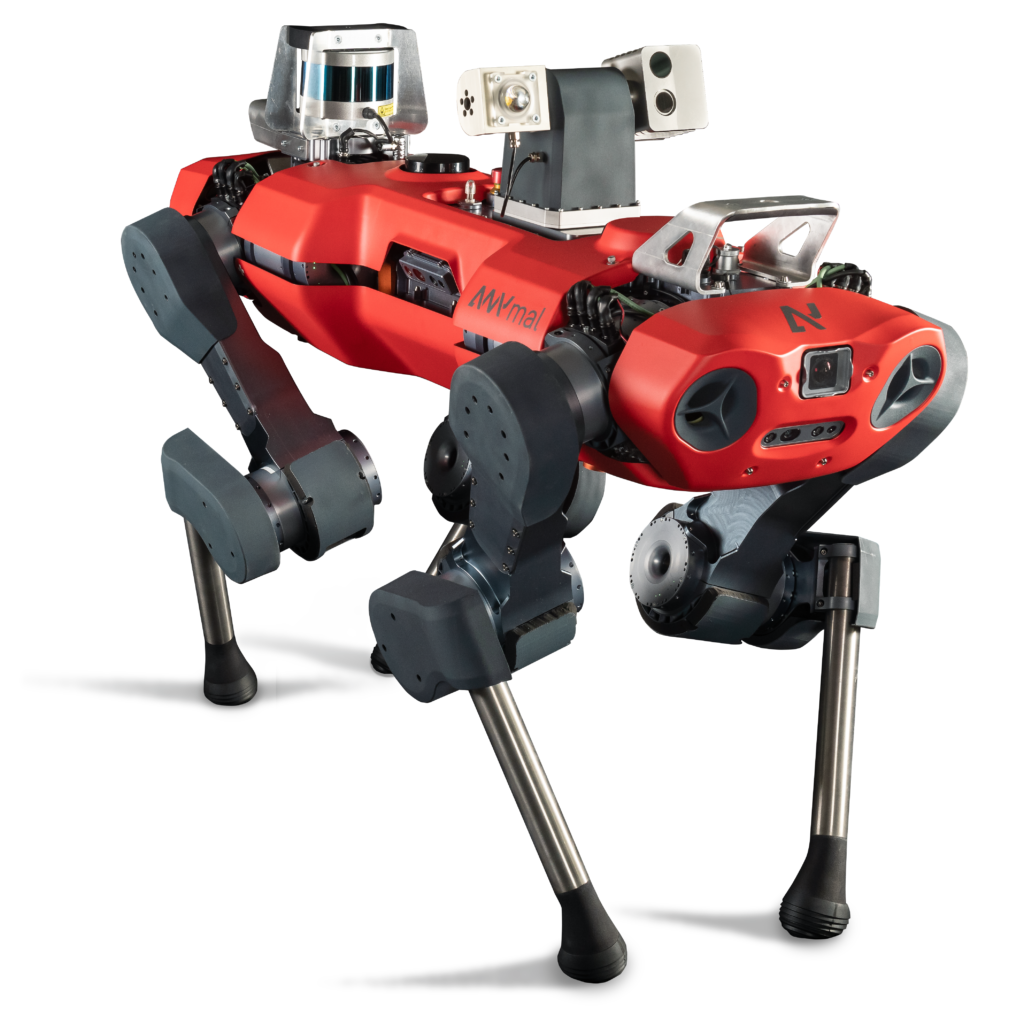}
    \end{subfigure}
    \begin{subfigure}[t]{.83\linewidth}
        \includegraphics[width=\linewidth]{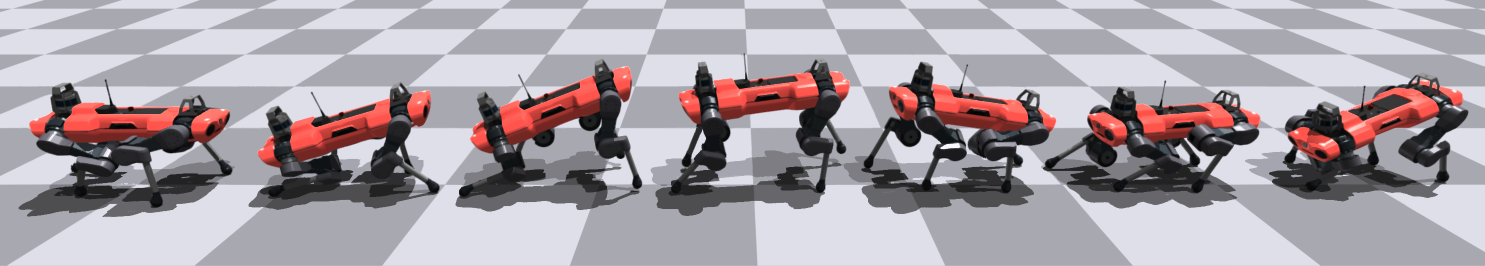}
    \end{subfigure}
	\caption{ANYmal C (left). Wave motion in Isaac Gym (right).}
    \label{fig:anymal}
    \end{figure}

    As the reference motion in \method{} contains only base information, it does not restrict itself to be obtained only from any specific robotic platform.
    This provides the possibility of cross-platform imitation.
    Using the reference trajectories recorded from \solo{}, with a manual offset of $0.25$\, m on the base height dimension addressing different sizes of the robots, we apply \method{} to ANYmal~\citep{anybotics}, a four-legged dog-like robot for research and industrial maintenance (\figref{fig:anymal}).
    To confirm that \method{} applies to cross-platform imitation, we define \anymalwave{} and \anymalbf{} tasks for the corresponding wave and backflip motions learned by ANYmal, yet from the reference data recorded from \solo{}.
    The performance in terms of the DTW distance is detailed in \tab{tab:anymal_performance}.

    \begin{table}[htb]
        \centering
        \begin{tabular}{lcccc} 
            \toprule
            Method  & \multicolumn{1}{c}{\solowave} & \multicolumn{1}{c}{\anymalwave}  & \multicolumn{1}{c}{\solobf}  &  \multicolumn{1}{c}{\anymalbf}   \\ 
            \midrule                                      
            \method{} &\boldmath$247.29\pm11.59$ & \boldmath$193.08\pm14.52$ & \boldmath $477.43\pm56.77$ &\boldmath $572.60\pm12.18$  \\
            \midrule
            Stand Still &  \multicolumn{2}{c}{$460.15$} & \multicolumn{2}{c}{$877.74$} \\
            \bottomrule
        \end{tabular}
        \vspace{0.5em}
        \caption{Performance of cross-platform imitation of ANYmal using \method{} trained with hand-held demonstrations from \solo in terms of \textbf{DTW distance} $\ddtw$, successful runs are in \textbf{bold} font.}
        \label{tab:anymal_performance}
    \end{table}
    

\section{Conclusion}
\label{sec:conclusion}

In this work, we propose an adversarial imitation method named \method{} for inferring reward functions that is capable of learning agile skills from partial and physically incompatible demonstrations without any a priori known reward terms. 
Our results indicate that \method{} allows extracting robust policies that are able to transfer to the real system and enables cross-platform imitation.
Furthermore, our experiments confirm that imitation learning using the \baseline{} fits style transfer settings where desired motions are more achievable.
For highly agile or incompatible motions which initially seem beyond the robot's capability, \method{} outperforms \baseline{} by successful and faithful replication of roughly demonstrated behaviors.
Further extensions and applications are presented in \supp{app:sec:extensions}.

\section{Limitations}
    
    While saving the effort of developing a specific task reward that motivates desired motions, providing a good evaluation metric in terms of a distance to the reference motion is not straightforward for generic rough demonstrations.
    Although DTW is a feasible option, it still requires a reasonable distance metric and careful choice of the warping procedure, which might be task-dependent. 
    Moreover, since our method works with rough demonstrations, even a good distance metric to the reference may not help inform about closeness to feasible, desirable motions from the robot's perspective.
    Finally, we do not intensively study to what extent our method is robust against the degree of incompatibility of the demonstrations.
    

\clearpage
\acknowledgments{
Georg Martius is a member of the Machine Learning Cluster of Excellence, EXC number 2064/1 – Project number 390727645.
We acknowledge the support from the German Federal Ministry of Education and Research (BMBF) through the Tübingen AI Center (FKZ: 01IS18039B). The authors thank the International Max Planck Research School for Intelligent Systems (IMPRS-IS) for supporting Marin Vlastelica and Sebastian Blaes, and Max Planck ETH Center for Learning Systems for supporting Jonas Frey.
}



\clearpage

\appendix
\appendix
\renewcommand{\thetable}{S\arabic{table}}
\renewcommand{\thefigure}{S\arabic{figure}}
\renewcommand{\theequation}{S\arabic{equation}}


\begin{center}
    \Large \textbf{Supplementary for \papertitle{}}
\end{center}

\section{Training Details}\label{app:sec:training_details}

\subsection{Training Parameters}

    The learning networks and algorithm are implemented in PyTorch 1.10 with CUDA 11.3. Adam is used as the optimizer for the policy and value function with an adaptive learning rate with a KL divergence target of $0.01$. The optimizer for the discriminator is SGD for \baseline and RMSprop for \method. The discount factor $\gamma$ is set to $0.99$, the clip range $\epsilon$ is set to $0.2$, and the entropy coefficient $\alpha$ is set to $0.01$. The policy runs at $50$\,Hz. All training is done by collecting experiences from $4096$ uncorrelated instances of the simulator in parallel. Most of the experiments are executed on the cluster of Max Planck Institute for Intelligent Systems with NVIDIA A100 and Tesla V100 GPUs. In this setting, one run with $5000$ iterations with the specified compute settings and devices completes within $2$ hours. The information is summarized in \tab{table:training_params}.
    \begin{table}[h]
    \centering
        \caption{Training parameters}
        \begin{tabular}{lcc}
        \toprule
            Parameter & Symbol & Value \\
            \midrule
            step time seconds & $-$ & $0.02$ \\
            max episode time seconds & $-$ & $20$ \\
            max iterations & $-$ & $5000$ \\
            steps per iteration & $-$ & $24$ \\
            policy learning epochs & $-$ & $5$ \\
            policy mini-batches & $-$ & $4$ \\
            KL divergence target & $-$ & $0.01$ \\
            discount factor & $\gamma$ & $0.99$ \\
            clip range & $\epsilon$ & $0.2$ \\
            entropy coefficient & $\alpha$ & $0.01$ \\
            discriminator weight decay & $wd^\mathrm{D}$ & $0.001$ \\
            discriminator momentum & $-$ & $0.05$ \\
            discriminator gradient penalty coefficient & $-$ & $5.0$ \\
            discriminator learning epochs & $-$ & $1$ \\
            discriminator mini-batches & $-$ & $80$ \\
            Wasserstein loss weight & $\wD$ & $0.5$ \\
            gradient penalty weight & $\wGP$ & $5.0$ \\
            parallel training environments & $-$ & $4096$ \\
            number of seeds & $-$ & $5$ \\
            approximate training hours & $-$ & $2$ \\
            \midrule
            policy learning rate & $lr^\pi$ & grid-searched \\
            discriminator learning rate & $lr^\mathrm{D}$ & grid-searched \\
            discriminator observation horizon & $H$ & grid-searched \\
            imitation reward relative importance & $\wI$ & grid-searched \\
        \bottomrule
        \end{tabular}
        \label{table:training_params}
    \end{table}

    Note that policy learning rate, discriminator learning rate, discriminator observation horizon, and imitation reward relative importance are grid-searched and optimized in different tasks. We report the value used for evaluation in \sec{sec:experiments} in \tab{table:task_specific_params}.

    \begin{table}
        \centering
        \caption{Task-specific parameters}
        \begin{tabular}{lllcccc}
        \toprule
            Section & Task & Method& $lr^\pi$ & $lr^\mathrm{D}$ & $H$ & $\wI$ \\
            \midrule
            \sec{sec:wgan_vs_lsgan} & \solobf & \baseline & $1.0 \times 10^{-6}$ & $1.0 \times 10^{-4}$ & $2$ & $0.8$  \\
            (Sim) & & \method & $1.0 \times 10^{-7}$ & $1.0 \times 10^{-7}$ & $16$ & $4.0$ \\
            \midrule
            \sec{sec:learning_to_mimic_rough_demonstrations} & \sololeap & \baseline & $1.0 \times 10^{-4}$ & $1.0 \times 10^{-4}$ & $2$ & $0.8$ \\
            (Sim) & & \method & $1.0 \times 10^{-7}$ & $5.0\times10^{-8}$ & $2$ & $8.0$ \\
            & \solowave & \baseline & $1.0 \times 10^{-6}$ & $1.0 \times 10^{-4}$ & $4$ & $0.8$ \\
            & & \method & $1.0 \times 10^{-6}$ & $5.0 \times10^{-8}$ & $4$ & $0.8$ \\
            & \solostandup & \baseline & $1.0 \times 10^{-4}$ & $1.0 \times 10^{-4}$ & $2$ & $0.8$ \\
            & & \method & $1.0 \times 10^{-7}$ & $1.0 \times 10^{-7}$ & $4$ & $4.0$ \\
            & \solobf & \baseline & $1.0 \times 10^{-6}$ & $1.0 \times 10^{-4}$ & $2$ & $0.8$ \\
            & & \method &  $1.0 \times 10^{-7}$ & $1.0 \times 10^{-7}$ & $16$ & $4.0$ \\
            \cmidrule{2-7}
            & \solostanduphcp & \baseline & $5.0 \times 10^{-5}$ & $5.0 \times 10^{-5}$ & $8$ & $0.8$ \\
            & & \method & $1.0 \times 10^{-7}$ & $5.0 \times 10^{-7}$ & $2$ & $4.0$ \\
            & \solostanduphcf & \baseline & $1.0 \times 10^{-4}$ & $1.0 \times 10^{-5}$ & $8$ & $0.8$ \\
            & & \method & $5.0 \times 10^{-7}$ & $1.0 \times 10^{-7}$ & $2$ & $4.0$ \\
            & \solobfhcp & \baseline & $5.0 \times 10^{-5}$ & $1.0 \times 10^{-5}$ & $8$ & $0.8$ \\
            & & \method & $1.0 \times 10^{-7}$ & $5.0 \times 10^{-8}$ & $2$ & $4.0$ \\
            & \solobfhcf & \baseline & $1.0 \times 10^{-4}$ & $1.0 \times 10^{-5}$ & $8$ & $0.8$ \\
            & & \method & $1.0 \times 10^{-7}$ & $5.0 \times 10^{-8}$ & $2$ & $0.8$ \\
            \midrule
            \sec{sec:evaluation_on_real_robot} & \sololeap & \method & $1.0 \times 10^{-3}$ & $5.0 \times 10^{-7}$ & $2$ & $0.1$ \\
            (Real) & \solowave & \method & $1.0 \times 10^{-3}$ & $5.0 \times 10^{-7}$ & $2$ & $0.1$ \\
            & \solobf & \method & $1.0 \times 10^{-3}$ & $5.0 \times 10^{-7}$ & $2$ & $0.4$ \\
        \bottomrule
        \end{tabular}
        \label{table:task_specific_params}
    \end{table}

\subsection{Network Architecture}

The network architecture is detailed in \tab{table:network_architecture}, where $H$ denotes the discriminator observation horizon.

\begin{table}
    \centering
    \caption{Network architecture}
    \begin{tabular}{lcccc}
    \toprule
        Network & Symbol & Type & Shape & Activation \\
        \midrule
        policy & $\pi$ & MLP & $68, 128, 128, 128, 8$ & Exponential Linear Unit (ELU) \\
        value function & $V$ & MLP & $68, 128, 128, 128, 1$ & Exponential Linear Unit (ELU) \\
        discriminator & $D$ & MLP & $10 H, 512, 256, 1$ & Rectified Linear Unit (ReLU) \\
    \bottomrule
    \end{tabular}
    \label{table:network_architecture}
\end{table}

\subsection{Domain Randomization}

    Two types of domain randomization techniques are applied during training to improve policy performance when transferring from simulation to the real system.

    On the one hand, the base mass of the parallel training instances is perturbed with an additional weight $m' \sim \mathcal{U}(-0.5, 1.0)$, where $\mathcal{U}$ denotes uniform distribution. On the other hand, random pushing is also applied every $15$ seconds on the robot base by forcing its horizontal linear velocity to be set randomly within $v_{xy} \sim \mathcal{U}(-0.5, 0.5)$.

\section{Model Representation} \label{app:sec:model_representation}

\subsection{Discriminator Observation} \label{app:sec:discriminator_observation}

    \tab{table:discriminator_observation_space} lists the extracted features sent to the discriminator.

    \begin{table}
        \centering
        \caption{Discriminator observation space}
        \begin{tabular}{lcc}
        \toprule
            Entry & Symbol & Dimensions \\
            \midrule
            base linear velocity & $v$ & 0:3 \\
            base angular velocity & $\omega$ & 3:6 \\
            projected gravity & $g$ & 6:9 \\
            base height & $z$ & 9:10 \\
        \bottomrule
        \end{tabular}
        \label{table:discriminator_observation_space}
    \end{table}

    For discriminator observation horizon $H > 1$, the entries are concatenated to a vector of size $10H$.
    The resulting features are then normalized and used as inputs to the discriminator network.

\subsection{Policy Observation and Action Space} \label{app:sec:policy_observation}
    In our work, we use a universal set of states as the policy observations for all tasks.
    It has 68 dimensions and consists of the same set of state measurements of two consecutive steps as detailed in \tab{table:policy_observation_space}.
    The observation space is composed of base linear and angular velocities $v$, $\omega$ in the robot frame, measurement of the gravity vector in the robot frame $g$, base height $z$, joint positions $q$ and velocities $\dot{q}$, and the most recent actions $a'$.

    \begin{table}
        \centering
        \caption{Policy observation space}
        \begin{tabular}{lccc}
        \toprule
            Entry & Symbol & Dimensions & noise level $b$ \\
            \midrule
            base linear velocity & $v$ & 0:3 & $0.2$ \\
            base angular velocity & $\omega$ & 3:6 & $0.05$ \\
            projected gravity & $g$ & 6:9 & $0.05$ \\
            base height & $z$ & 9:10 & $0.01$ \\
            joint positions & $q$ & 10:18 & $0.01$ \\
            joint velocities & $\dot{q}$ & 18:26 & $0.75$ \\
            last actions & $a'$ & 26:34 & $0.0$ \\
        \bottomrule
        \end{tabular}
        \label{table:policy_observation_space}
    \end{table}

    The noise level $b$ denotes the artificial noise added during training to increase the policy robustness.
    Note again that the policy receives the observation collection for two consecutive steps.
    This is not necessarily optimal for all the tasks as less information would suffice for easier tasks.
    For simplicity, we use this fixed set of observations for all experiments.

    The action space is of 8 dimensions and encodes the target joint position for each of the 8 actuators. The PD gains are set to $5.0$ and $0.1$, respectively.

\section{Regularization Reward Functions} \label{app:sec:regularization_reward_functions}

    The regularization reward functions contain only regularization terms whose formulation is detailed below. A universal set of involved hyperparameters is used across different motions and is summarized in \tab{table:regularization_reward_hyperparameters}.

    \begin{table}
    \centering
        \caption{Regularization reward hyperparameters}
        \begin{tabular}{lcccccc}
        \toprule
            Hyperparameter & $w_{ar}$ & $w_{q_a}$ & $w_{q_T}$ & $w_{\dot{\phi}}$ & $w_{\dot{\psi}}$ & $w_{\dot{y}}$ \\
            \midrule
            Value & $-0.005$ & $-1.25 \times 10^{-8}$ & $-1.25 \times 10^{-6}$ & $-0.001$ & $-0.001$ & $-0.001$ \\
            \bottomrule
        \end{tabular}
        \label{table:regularization_reward_hyperparameters}
    \end{table}

\subsection{Action Rate}

    \begin{equation}
        r_{ar} = w_{ar} \| a' - a \|_2^2,
    \end{equation}

    where $w_{ar}$ denotes the weight of the action rate reward, $a'$ and $a$ denote the previous and current actions.

\subsection{Joint Acceleration}

    \begin{equation}
        r_{q_a} = w_{q_a} \left \| \dfrac{\dot{q}' - \dot{q}}{\Delta t} \right \|_2^2,
    \end{equation}

    where $w_{q_a}$ denotes the weight of the joint acceleration reward, $\dot{q}'$ and $\dot{q}$ denote the previous and current joint velocity, $\Delta t$ denotes the step time interval.

\subsection{Joint Torque}

    \begin{equation}
        r_{q_T} = w_{q_T} \left \| T \right \|_2^2,
    \end{equation}

    where $w_{q_T}$ denotes the weight of the joint torque reward, $T$ denotes the joint torques.

\subsection{Angular Velocity $x$}

    \begin{equation}
        r_{\dot{\phi}} = w_{\dot{\phi}} \left \| \dot{\phi} \right \|_2^2,
    \end{equation}

    where $w_{\dot{\phi}}$ denotes the weight of the angular velocity $x$ reward, $\dot{\phi}$ denotes the base roll rate.

\subsection{Angular Velocity $z$}

    \begin{equation}
        r_{\dot{\psi}} = w_{\dot{\psi}} \left \| \dot{\psi} \right \|_2^2,
    \end{equation}

    where $w_{\dot{\psi}}$ denotes the weight of the angular velocity $z$ reward, $\dot{\psi}$ denotes the base yaw rate.

\subsection{Linear Velocity $y$}

    \begin{equation}
        r_{\dot{y}} = w_{\dot{y}} \left \| \dot{y} \right \|_2^2,
    \end{equation}

    where $w_{\dot{y}}$ denotes the weight of the linear velocity $y$ reward, $\dot{y}$ denotes the base lateral velocity.

\section{Method Comparison}\label{app:sec:method_comparison}

    In this section, we highlight the connections and differences between our work and the AMP work~\citep{peng2021amp_supp}, in particular, how the LSGAN formulation in our setting relates to AMP.

\subsection{AMP Adapted for Task Learning Termed as \baseline}

    It is acknowledged that both \method{} and AMP utilize generative adversarial learning methods to learn motions from reference demonstrations.
    However, from a high-level view, \method{} aims to solve fundamentally different tasks as opposed to AMP.
    
    Note that AMP itself does not target task reward learning. In AMP, the GAN model is employed to shape the styles of a learning agent while performing some other tasks, which are relatively straightforward and motivated by additional task reward functions (e.g. moving forward, reaching a target).
    This can be viewed as learning the regularization of motions.
    Indeed, we can adapt AMP to enable it to directly learn complex tasks by providing the desired motions as the reference and removing the original task reward.
    In addition, to alleviate the mode collapse issue as described in \sec{sec:preventing_mode_collapse}, the capability of the discriminator is extended to encompass longer observation horizons.
    With these adaptations, AMP is referred to as \baseline{} in our work.
    
    Typically, AMP learns demonstrated motions with sufficient exploration enabled by reference state initialization (RSI), where the agent is initialized randomly along the reference trajectories.
    However, in the setting of legged skill learning from only base information, RSI is not applicable due to the missing joint reference.
    This makes the robot difficult to directly imitate highly dynamic motions.
    And as shown in \tab{tab:performance}, the robot fails to adopt complex skills (e.g. \solostandup{}, \solobf{}) with the \baseline{} implementation.

\subsection{\method Designed for Task Learning}

    In contrast, \method{} is proposed to learn task rewards directly for agile motions from limited data.
    This becomes especially meaningful for tasks where the reward function is challenging to design and a decent expert is not immediately accessible.
    \method{} provides solutions to cases where we want to quickly develop a complex skill for robot learning, allowing us to hand-hold a robot (or an object) without actuating it and to learn directly from the demonstrated trajectories.

\section{Handcrafted Task Reward}\label{app:sec:handcrafted_task_reward}

    Experts trained on handcrafted task rewards are used as a baseline to prove that \method{} is capable of extracting sensible task rewards. The policies learned using LSGAN and \method{} use 1000 sampled trajectories generated by the experts in simulation as reference.

\subsection{Upright Stand for \solostandup{}}

    The robot is encouraged to stand up by rewarding the pitch angle of the base, the base height, and the standing on only the two hind legs.

    \begin{equation}
        r_{\beta} = w_{\theta_z} \theta_z + w_{z} z + w_{g} c_g,
    \end{equation}

    where $\theta_z$ denotes the pitch angle with respect to the global $z$-axis, $w_{\theta_z} = 1.0$ denotes its weight. $z$ denotes the base height and $w_z = 3.0$ denotes its weight. $c_g$ is a binary variable that takes $1$ if the front legs have no contact with the ground, $w_{g} = 2.0$ denotes its weight.

\subsection{Traversed Angle for \solobf{}}

    The robot is encouraged to perform back-flipping by rewarding its traversed angle around the $y$-axis while in the air. This reward will be given only when the robot lands.

    \begin{equation}
        r_{\theta} = w_{\theta} \theta \llbracket s\in \gL \rrbracket,
    \end{equation}

    where $w_{\theta} = 5.0$ denotes the weight of the traversed angle reward, $\theta$ denotes the traversed angle around $y$-axis while in the air. $\gL$ is the set of robot landing states, and $\llbracket \cdot \rrbracket$ is the Iverson bracket.

\section{Sim-to-Real Transfer}\label{app:sec:sim-to-real_transfer}

    To deploy the learned policy on the real system, the following adaptations are made to reduce the sim-to-real gap.

    \subsection{Observation Space Adaptation}

    Generally, policies trained in simulation suffer from poor performance when transferred to the real system. One primary reason for such failures is the incorrectly modeled system dynamics. As a result, the state transitions observed in simulation may divert from reality. On account of this, instead of using two consecutive steps of the observation collection in \supp{app:sec:policy_observation}, only the observation at the current time step is used. The resulting policy observation space has thus only 34 entries and is detailed in \tab{table:policy_observation_space}.

    \subsection{Training Hyperparameter Adaptation}

    Some training hyperparameters are further refined and specifically adapted for the deployment on the real system as detailed in \tab{table:task_specific_params}.

    \subsection{Reward Adaptation}

    To generate stable joint actuation and achieve task-specific high performance, a feet air time regularization reward is introduced to motivate higher off-ground steps during robot movement.

    \begin{equation}
        r_{t_f} = w_{t_f} \sum_{i=1}^4 t_{f_i} \llbracket f_i\in \gC \rrbracket,
    \end{equation}

    where $w_{t_f}$ denotes the weight of the feet air time reward, $t_{f_i}$ denotes the time foot $i$ stays in the air. $\gC$ is the set of foot states touching the ground, and $\llbracket \cdot \rrbracket$ is the Iverson bracket.
    The resulting task-specific regularization reward setting is provided in \tab{table:regularization_reward_adaptation}.

    \begin{table}
    \centering
        \caption{Regularization reward adaptation}
        \begin{tabular}{lccccccc}
        \toprule
            Task & $w_{ar}$ & $w_{q_a}$ & $w_{q_T}$ & $w_{\dot{\phi}}$ & $w_{\dot{\psi}}$ & $w_{\dot{y}}$ & $w_{t_f}$\\
            \midrule
            \sololeap & $-0.01$ & $-2.5 \times 10^{-7}$ & $-2.5 \times 10^{-5}$ & $-0.05$ & $-0.05$ & $-0.05$ & $0.10$ \\
            \solowave & $-0.01$ & $-2.5 \times 10^{-7}$ & $-2.5 \times 10^{-5}$ & $-0.01$ & $-0.01$ & $-0.01$ & $0.05 $\\
            \solobf & $-0.01$ & $-2.5 \times 10^{-7}$ & $-1.0 \times 10^{-5}$ & $-0.02$ & $-0.02$ & $-0.02$ & $0.00$\\
            \bottomrule
        \end{tabular}
        \label{table:regularization_reward_adaptation}
    \end{table}



    \subsection{Policy Transfer}

    In our work, the learned policies for \sololeap{}, \solowave{}, and \solobf{} successfully transfer to the real system.
    However, during the deployment of the policy for \solostandup{} from either the handcrafted expert or obtained by \method{}, the robot fails to maintain balance after it manages to stand up.
    As \solo does not have Hip Abduction Adduction (HAA) joints, it could be challenging to adapt against unexpected base rolling accordingly, which happens frequently after standing up.
    This makes this task especially difficult.
    In addition, the sim-to-real gap could also be potentially encoded in the inaccurate modeling of the point contacts between the feet and the ground, which the robot requires to adapt constantly to maintain its attitude.

\section{Motions}\label{app:sec:tasks}

\subsection{Data Collection}

\begin{figure}
    \centering
        \includegraphics[width=1.0\linewidth]{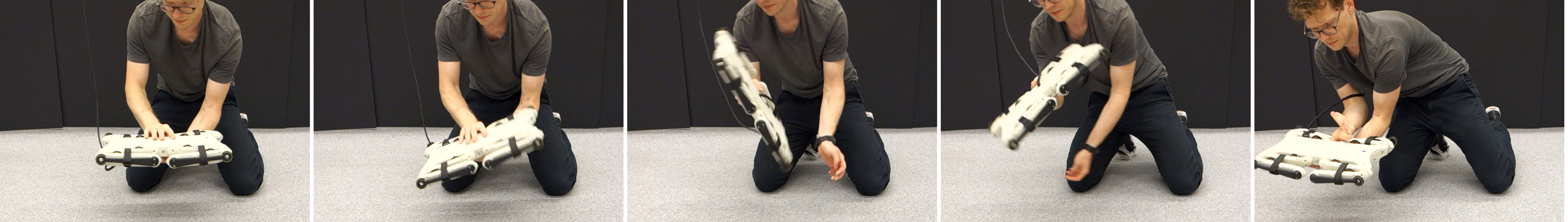}
    \caption{Hand-held motion demonstration for \solobf. Note that the robot joints are not actuated and only base information is recorded.}
    \label{fig:data_collection}
    \end{figure}

    The reference motions containing only the base information (as detailed in \sec{app:sec:discriminator_observation}) are performed by a human demonstrator and recorded using a Vicon motion capture system as illustrated in \Fig{fig:data_collection}. For each motion, 20 trajectories are recorded and used as the reference motion dataset. The number of frames in the recorded trajectories for each motion is detailed in \tab{table:number_of_frames}.

    \begin{table}
        \centering
        \caption{Number of frames per recorded trajectory}
        \begin{tabular}{lcccc} 
            \toprule
            & \sololeap & \solowave & \solostandup & \solobf \\ 
            \midrule
            Frames  & $130$  & $130$ & $100$ & $60$ \\ 
            \bottomrule
        \end{tabular}
        \label{table:number_of_frames}
    \end{table}

\subsection{Motion Details}

\begin{figure}
    \centering
    \begin{subfigure}[b]{1.0\linewidth}
    \centering
        \includegraphics[width=1.0\linewidth]{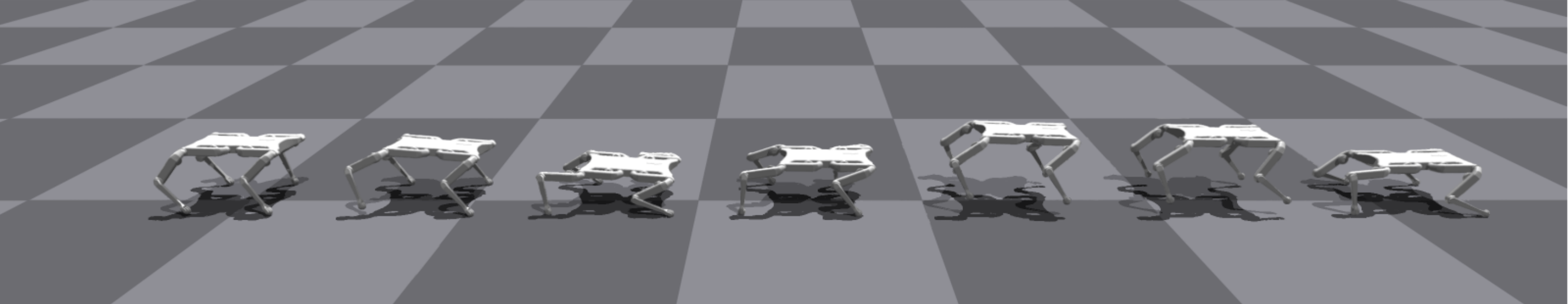}
        \caption{\sololeap}
    \end{subfigure}\hspace{0.1em}
    \begin{subfigure}[b]{1.0\linewidth}
    \centering
        \includegraphics[width=1.0\linewidth]{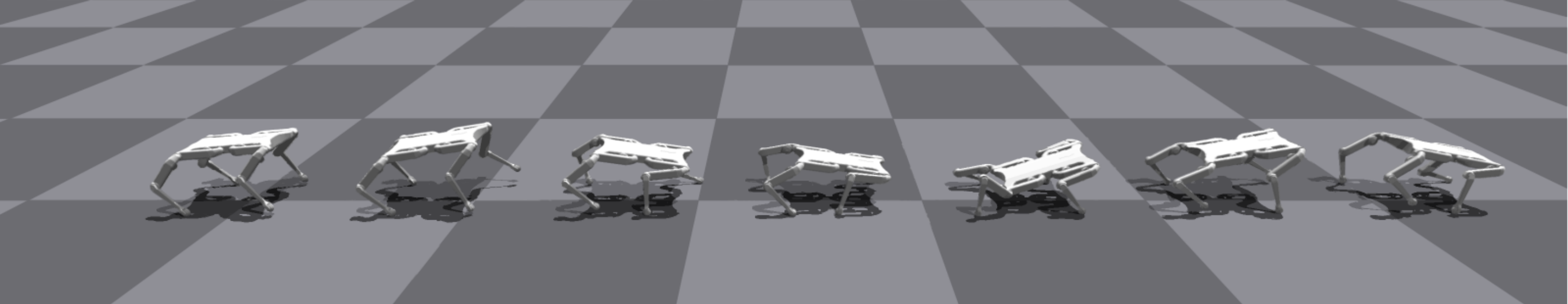}
        \caption{\solowave}
    \end{subfigure}\hspace{0.1em}
    \begin{subfigure}[b]{1.0\linewidth}
    \centering
        \includegraphics[width=1.0\linewidth]{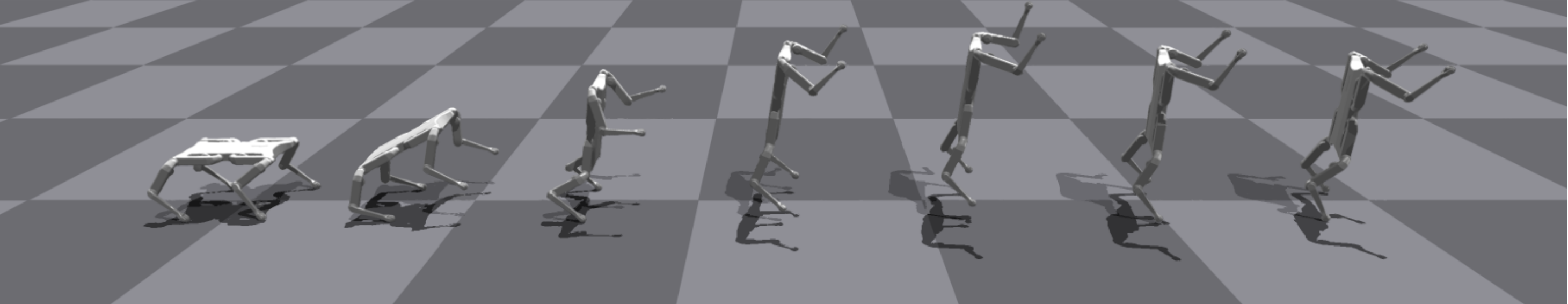}
        \caption{\solostandup}
    \end{subfigure}\hspace{0.1em}
    \begin{subfigure}[b]{1.0\linewidth}
    \centering
        \includegraphics[width=1.0\linewidth]{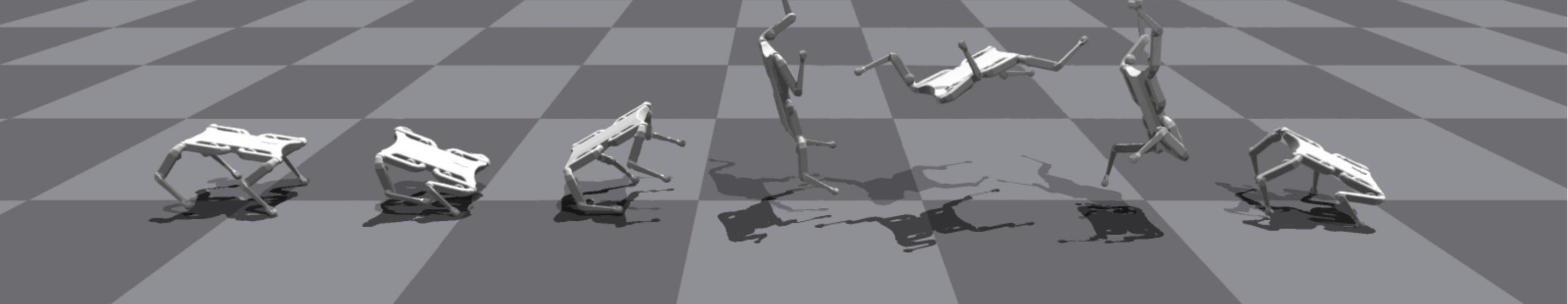}
        \caption{\solobf}
    \end{subfigure}\hspace{0.1em}
    \caption{Task motion sequences induced by the learned policies in simulation.}
    \label{fig:motions}
\end{figure}

We provide sequences of the respective motions that we learn in this work in \fig{fig:motions}.

\section{Dynamic Time Warping Evaluation} \label{app:sec:dtw}

We make use of an imitation reward as outlined in \sec{sec:wgan} for learning policies from rough reference trajectories.
Since the reference trajectories might not be completely achievable, either in terms of time consistency or the actual sequence of states that are infeasible, a simple evaluation metric such as the direct $L_2$ distance between policy and reference trajectories is not applicable.
Furthermore, a learned policy might replicate the demonstrated motion perfectly in a certain sub-sequence, however, in the absence of proper alignment and synchronization, it would incur a large penalty in terms of a simple distance metric.

To account for the potential misalignment in the policy vs. reference trajectories, the policy trajectories need to be synchronized to the reference.
Dynamic Time Warping (DTW) is an algorithm that computes an alignment between two sequences such that a distance measure between the elements of the sequences is minimized, under a set of matching constraints.
This can be computed efficiently via dynamic programming, an example of such a matching is given for our case in \figref{fig:demo-distance}.

For the distance measure between the matched elements of the sequences we use the standard $L_2$ distance, and for the matching computation, we use the `dtw` Python package~\citep{giorgino2009computing} with the Mori asymmetric step pattern~\citep{mori2006early} and open-ended matching. The asymmetric step pattern constrains the type of element matches that can happen between the trajectories and also makes it possible for some elements to be skipped, while open-ended matching allows for ends of trajectories to be matched to an earlier time point which is necessary to account for time shift. The resulting measure is denoted with $\ddtw$.

As clarified in \sec{sec:experiments},  we calculate the expectation
\begin{equation}
    \E\left[\ddtw(\Phi(\traj_\pi), \traj_\gM)\right]
\end{equation}
as the evaluation metric, where $\traj_\pi\sim d^\pi$ is a state trajectory drawn from a policy rollout distribution and $\traj_\gM\sim d^\gM$ denotes a reference motion from the dataset.
With the same notation as in \sec{sec:preventing_mode_collapse}, the function $\Phi$ maps each state in the state trajectory of the policy to the reference observation space $\mathcal{O}$.
In practice, we estimate this by drawing 20 policy rollouts, comparing each of the 20 rollouts with 20 collected reference trajectories, and taking the mean.
Note that $\ddtw$ is not comparable across different tasks, thus we provide a pure standing reference in \tab{tab:performance}.

\section{Ablation Studies} \label{app:sec:ablation}

\begin{figure}
    \centering
    \vspace{1em}
    \begin{subfigure}[t]{.36\linewidth}
        \ \\[-1em]
        \includegraphics[width=\linewidth]{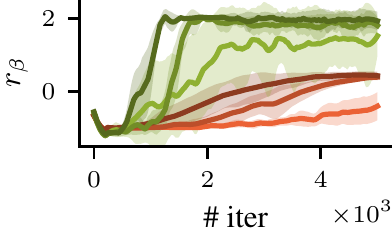}
    \end{subfigure}
    \hspace{1em}
    \begin{subfigure}[t]{.36\linewidth}
        \ \\[-1em]
        \includegraphics[width=\linewidth]{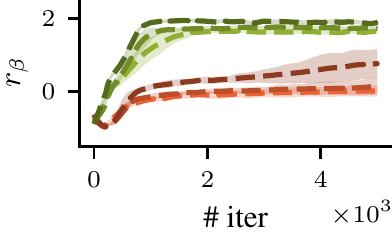}
    \end{subfigure}
    \hspace{1em}
    \begin{subfigure}[t]{.2\linewidth}
        \ \\
        \ourlegendvert
    \end{subfigure}

    \caption{Performance of \baseline{} (left) and \method{} (right) in terms of the handcrafted task reward for \solostandup{} with different discriminator observation horizons (light $H=2$, middle $H=4$, dark $H=8$). Solid lines indicate full information ($*$) and dashed lines indicate partial information ($\dagger$).}
    \label{fig:ablation}
    \end{figure}

    To prevent mode collapse, we extend the capability of the discriminator by allowing more than one state transition as input. In this section, we aim to investigate how the length of the discriminator horizon may affect the learning process of the desired behaviors. Here we provide an evaluation in terms of the handcrafted reward of policies learned with discriminator observations of horizon $H = 2, 4, 8$ with \baseline and \method in \solostandup. The result is depicted in \fig{fig:ablation}.

Observe that a longer horizon tends to help the policy converge earlier and yield slightly better performance in terms of the handcrafted reward in both methods with either full or partial reference information, even if the \baseline fails to learn the stand-up behavior.
A similar pattern is also revealed in \solobf, although it presents a weaker effect on policy convergence in comparison.
The potential reason is that the policy learns to avoid producing state transitions that align with only a short sub-sequence of the reference motion.
Such avoidance of mode collapse would prevent the policy from getting stuck at the local optima and thus increase the overall performance.

However, the benefit brought about by a longer discriminator observation horizon is not identified in tasks with hand-held reference motions where a direct performance metric is not applicable.
Sometimes a longer horizon may even result in failure of learning the desired motion which is attainable with shorter horizons.
An evaluation of performance improvement in terms of DTW over iterations reveals no clear pattern on how different discriminator observation horizons affect policy convergence.
This may result from the time and state inconsistency in the hand-held reference motions which may already alleviate mode collapse to some extent.

\section{Extensions} \label{app:sec:extensions}

    In this section, we present extensions of \method{} by small modifications to the adversarial imitation learning framework proposed in our work.
    Supplementary videos for these extensions are available at \url{https://sites.google.com/view/corl2022-wasabi/home}.

    \subsection{Active Velocity Control}
    
    In our work, the hand-held reference motions are recorded by a human demonstrator.
    For locomotion tasks, if the demonstration speed is uniformly high in the reference dataset, the corresponding recorded velocity terms are also high.
    During the training of the policy, the robot will try to mimic the recorded data and thus operate at a high locomotion speed.

    In contrast, if there is some variance in the velocity terms within the reference dataset, the policy will have larger freedom in choosing its locomotion velocity.
    For this reason, we take advantage of the variance within the human demonstration and realize active velocity control by introducing a target-conditioned velocity tracking reward as follows
    
    \begin{equation}
        r_{\dot{x}} = w_{\dot{x}} e^{-\dfrac{(c_{\dot{x}} - \dot{x})^2}{\sigma^2_{\dot{x}}}},
    \end{equation}

    where $w_{\dot{x}} = 0.5$ denotes the weight of the velocity tracking reward, $\dot{x}$ denotes the base longitudinal velocity, $c_{\dot{x}}$ denotes the velocity command on the same direction, which is also observed by the control policy. And $\sigma^2_{\dot{x}} = 0.25$ denotes a temperature scale for the tracking error.
    
    For \sololeap{} and \solowave{}, we successfully enable active velocity control within the range $c_{\dot{x}} \in [0.1, 0.5]$ by utilizing the diversity of the demonstration speed within the reference dataset.
    
    \subsection{Single Flip in \solobf}
    
    Using the same reference motion in \solobf{}, active execution of a single flip is achieved by including an additional variable in the observation space of the control policy indicating whether the robot has finished a backflip.
    After the robot finishes a flip, the control policy stops receiving imitation signals from the discriminator by fixing a constant imitation reward output to prevent from learning a consecutive flip.

    In the meanwhile, a stand-still reward term is imposed to regularize the joint configurations during landing and standing.
    The stand-still reward is formulated as follows
    
    \begin{equation}
        r_{q_0} = w_{q_0} \left \| q - q_0  \right \|_2^2,
    \end{equation}
    
    where $w_{q_0} = 0.01$ denotes the weight of the stand-still reward, $q$ denotes the current joint positions and $q_0$ denotes the default joint positions when the robot stands.

\clearpage


\end{document}